\documentclass[11pt,a4paper]{article}
\usepackage[hyperref]{acl2019}
\usepackage{times}
\usepackage{latexsym}
\usepackage{url}

\usepackage{algorithm}
\usepackage{algorithmic}
\usepackage{enumerate}
\usepackage{amsmath} 
\usepackage{amssymb}  

\usepackage{booktabs}
\usepackage{multirow}
\usepackage{float}
\usepackage{graphicx}
\usepackage{footnote}
\usepackage{tablefootnote}
\usepackage{subcaption}
\usepackage{breqn}
\usepackage{makecell}

\aclfinalcopy 
\def\aclpaperid{551} 

\newcommand*{\affaddr}[1]{#1} 
\newcommand*{\affmark}[1][*]{\textsuperscript{#1}}
\newcommand*{\email}[1]{\texttt{#1}}

\title{An Interactive Multi-Task Learning Network for End-to-End Aspect-Based Sentiment Analysis}

\author{Ruidan He\affmark[\dag\ddag], Wee Sun Lee\affmark[\dag], Hwee Tou Ng\affmark[\dag], \and Daniel Dahlmeier\affmark[\ddag]\\
\affaddr{\affmark[\dag]Department of Computer Science, National University of Singapore}\\
\affaddr{\affmark[\ddag]SAP Innovation Center Singapore}\\
\email{\affmark[\dag]\{ruidanhe,leews,nght\}@comp.nus.edu.sg}\\
\email{\affmark[\ddag]d.dahlmeier@sap.com}%
}

\date{}

\begin{document}
\maketitle
\begin{abstract}
Aspect-based sentiment analysis produces a list of aspect terms and their corresponding sentiments for a natural language sentence. This task is usually done in a pipeline manner, with aspect term extraction performed first, followed by sentiment predictions toward the extracted aspect terms. While easier to develop, such an approach does not fully exploit joint information from the two subtasks and does not use all available sources of training information that might be helpful, such as document-level labeled sentiment corpus. In this paper, we propose an interactive multi-task learning network (IMN) which is able to jointly learn multiple related tasks simultaneously at both the token level as well as the document level. Unlike conventional multi-task learning methods that rely on learning common features for the different tasks, IMN introduces a message passing architecture where information is iteratively passed to different tasks through a shared set of latent variables. Experimental results demonstrate superior performance of the proposed method against multiple baselines on three benchmark datasets. 
\end{abstract}

\section{Introduction}
Aspect-based sentiment analysis (ABSA) aims to determine people's attitude towards specific aspects in a review. This is done by extracting explicit aspect mentions, referred to as aspect term extraction (AE), and detecting the sentiment orientation towards each extracted aspect term, referred to as aspect-level sentiment classification (AS). For example, in the sentence \emph{``\underline{Great} \textbf{food} but the \textbf{service} is \underline{dreadful}''}, the aspect terms are \emph{``food''} and \emph{``service''}, and the sentiment orientations towards them are positive and negative respectively.  

In previous works, AE and AS are typically treated separately and the overall task is performed in a pipeline manner, which may not fully exploit the joint information between the two tasks. Recently, two studies~\citep{Wang:18b, Li:19} have shown that integrated models can achieve comparable results to pipeline methods. Both works formulate the problem as a single sequence labeling task with a unified tagging scheme\footnote{\{B, I\}-\{POS, NEG, NEU\} denotes the beginning and inside of an aspect-term with positive, negative, or neutral sentiment, respectively, and O denotes background words.}.
However, in their methods, the two tasks are only linked through unified tags, while the correlation between them is not explicitly modeled. Furthermore, the methods only learn from aspect-level instances, the size of which is usually small, and do not exploit available information from other sources such as related document-level labeled sentiment corpora, which contain useful sentiment-related linguistic knowledge and are much easier to obtain in practice.

In this work, we propose an interactive multi-task learning network (IMN), which solves both tasks simultaneously, enabling the interactions between both tasks to be better exploited. 
Furthermore, IMN allows AE and AS to be trained together with related document-level tasks, exploiting the knowledge from larger document-level corpora. 
IMN introduces a novel message passing mechanism that allows informative interactions between tasks.  Specifically, it sends useful information from different tasks back to a shared latent representation. The information is then combined with the shared latent representation and made available to all tasks for further processing. This operation is performed iteratively, allowing the information to be modified and propagated across multiple links as the number of iterations increases. 
In contrast to most multi-task learning schemes which share information through learning a common feature representation, IMN not only allows shared features, but also explicitly models the interactions between tasks through the message passing mechanism, allowing different tasks to better influence each other. 

In addition, IMN allows fined-grained token-level classification tasks to be trained together with document-level classification tasks. We incorporated two document-level classification tasks -- sentiment classification (DS) and domain classification (DD) --  to be jointly trained with AE and AS, allowing the aspect-level tasks to benefit from document-level information. 
In our experiments, we show that the proposed method is able to outperform multiple pipeline and integrated baselines on three benchmark datasets\footnote{Our source code can be obtained from \url{https://github.com/ruidan/IMN-E2E-ABSA}}.

\section{Related Work}\label{related work}

\noindent \textbf{Aspect-Based Sentiment Analysis.}
Existing approaches typically decompose ABSA into two subtasks, and solve them in a pipeline setting. Both AE~\citep{Qiu:11, Yin:16, Wang:16b, Wang:17, Li:17, He:17, Li:18b, stefanos18} and AS~\citep{Dong:14, Nguyen:15, Vo:15, Tang:16a, Wang:16, zhang:16, Liu:17, Chen:17, Cheng:17, Tay:18, Ma:18, He:18b, He:18, Li:18a} have been extensively studied in the literature. However, treating each task independently has several disadvantages. In a pipeline setting, errors from the first step tend to be propagated to the second step, leading to a poorer overall performance. In addition, this approach is unable to exploit the commonalities and associations between tasks, which may help reduce the amount of training data required to train both tasks.

Some previous works have attempted to develop integrated solutions. \citet{zhang:15} proposed to model the problem as a sequence labeling task with a unified tagging scheme. However, their results were discouraging. Recently, two works~\citep{Wang:18b, Li:19} have shown some promising results in this direction with more sophisticated network structures. However, in their models, the two subtasks are still only linked through a unified tagging scheme, while the interactions between them are not explicitly modeled. To address this issue, a better network structure allowing further task interactions is needed. 
\medskip

\noindent \textbf{Multi-Task Learning.}
One straightforward approach to perform AE and AS simultaneously is multi-task learning, where one conventional framework is to employ a shared network and two task-specific network to derive a shared feature space and two task-specific feature spaces. Multi-task learning frameworks have been employed successfully in various natural language processing (NLP) tasks~\citep{Collobert:08, Luong:15a, Liu:16}. By learning semantically related tasks in parallel using a shared representation, multi-task learning could capture the correlations between tasks and improve the model generalization ability in certain cases. For ABSA, \citet{He:18} have shown that aspect-level sentiment classification can be significantly improved through joint training with document-level sentiment classification. However, conventional multi-task learning still does not explicitly model the interactions between tasks -- the two tasks only interact with each other through error back-propoagation to contribute to the learned features and such implicit interactions are not controllable. Unlike existing methods, our proposed IMN not only allows the representations to be shared, but also explicitly models the interactions between tasks, by using an iterative message passing scheme. The propagated information contributes to both learning and inference to boost the overall performance of ABSA.
\medskip

\begin{figure*}[ht] 
\centering
\includegraphics[width=0.99\textwidth]{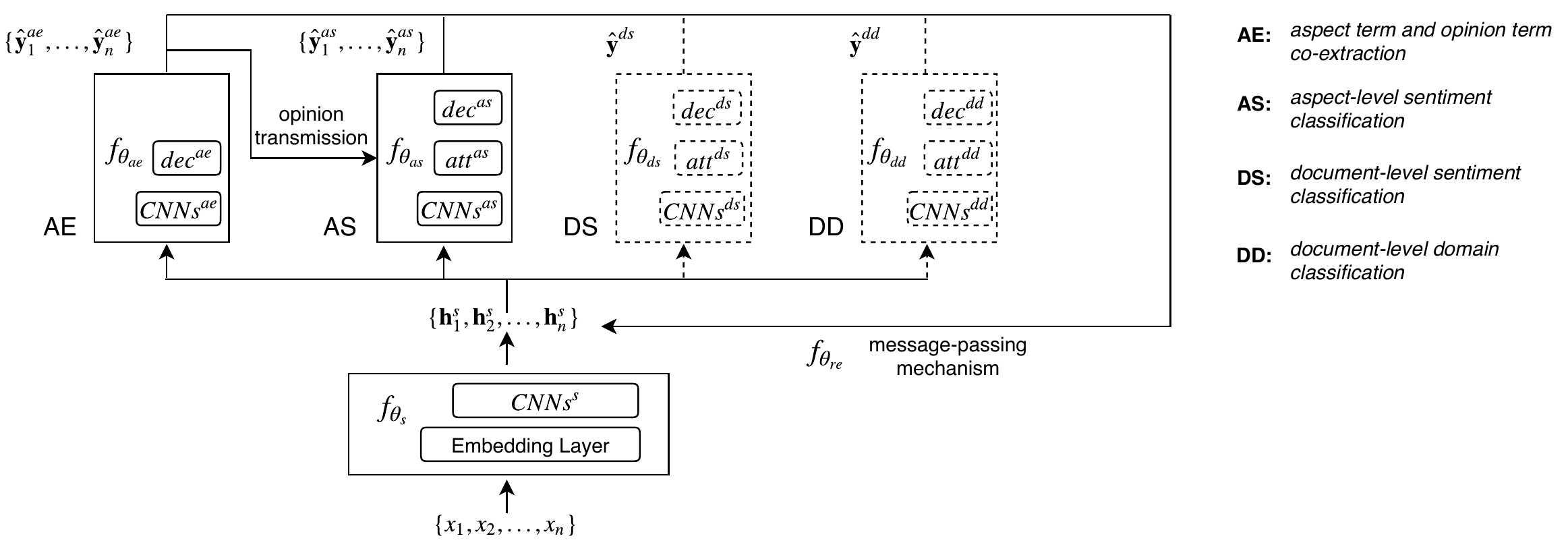}
\caption{The overall architecture of IMN.}
\label{overall_architecture_diagram}
\end{figure*}

\noindent \textbf{Message Passing Architectures.}
Networked representations for message passing graphical model inference algorithms have been studied in computer vision~\citep{arnab2018conditional} and NLP~\citep{gormley2015approximation}. Modeling the execution of these message passing algorithms as a network results in recurrent neural network architectures. 
We similarly propagate information in a network and learn the update operators, but the architecture is designed for solving multi-task learning problems. Our algorithm can similarly be viewed as a recurrent neural network since each iteration uses the same network to update the shared latent variables.

\section{Proposed Method}
The IMN architecture is shown in Figure~\ref{overall_architecture_diagram}. It accepts a sequence of tokens $\{x_1,\ldots,x_n\}$ as input into a feature extraction component $f_{\theta_s}$ that is shared among all tasks. This component consists of a word embedding layer followed by a few feature extraction layers. Specifically, we employ $m^s$ layers of CNNs after the word embedding layer in $f_{\theta_s}$.

The output of $f_{\theta_s}$ is a sequence of latent vectors $\{\mathbf{h}^s_{1}, \mathbf{h}^s_{2}, ..., \mathbf{h}^s_{n}\}$ shared among all the tasks. After initialization by $f_{\theta_s}$, this sequence of latent vectors is later updated by combining information propagated from different task components through message passing. We denote $\mathbf{h}^{s(t)}_i$ as the value of the shared latent vector corresponding to $x_i$ after $t$ rounds of message passing, with $\mathbf{h}^{s(0)}_i$ denoting the value after initialization. 

The sequence of shared latent vectors\footnote{We omit the iteration superscript $t$ in the description for simplicity.} $\{\mathbf{h}^s_{1}, \mathbf{h}^s_{2}, ..., \mathbf{h}^s_{n}\}$ is used as input to the different task-specific components. Each task-specific component has its own sets of latent and output variables. The output variables correspond to a label sequence in a sequence tagging task; in AE, we assign to each token a label indicating whether it belongs to any aspect or opinion\footnote{e.g. ``\emph{great}'' and ``dreadful'' in ``\emph{Great food but the service is dreadful}'' are the opinion terms.} term, while in AS, we label each word with its sentiment. 
In a classification task, the output corresponds to the label of the input instance: the sentiment of the document for the sentiment classification task (DS), and the domain of the document for the domain classification task (DD). At each iteration, appropriate information is passed back to the shared latent vectors to be combined; this could be the values of the output variables or the latent variables, depending on the task. In addition, we also allow messages to be passed between the components in each iteration. Specifically for this problem, we send information from the AE task to the AS task as shown in Figure~\ref{overall_architecture_diagram}. After $T$ iterations of message passing, which allows information to be propagated through multiple hops, we use the values of the output variables as predictions. For this problem, we only use the outputs for AE and AS during inference as these are the end-tasks, while the other tasks are only used for training.
We now describe each component and how it is used in learning and inference. 

\subsection{Aspect-Level Tasks}

\begin{table*}[t!]
\centering
\small
\begin{tabular}{lccccccccccccccc}
\toprule 
Input &The &fish &is &fresh &but &the &variety &of &fish &is &nothing &out &of &ordinary &.\\
AE &O &BA &O &BP &O &O &BA &IA &IA &O &O &O &O &BP &O\\
AS &- &\emph{pos} &- &- &- &- &\emph{neg} &\emph{neg} &\emph{neg} &- &- &- &- &- &-\\
\bottomrule
\end{tabular}
\caption{An aspect-level training instance with gold AE and AS labels.}\label{train_example}
\end{table*}

AE aims to extract all the aspect and opinion terms\footnote{Note that we are actually performing aspect and opinion term co-extraction. We still denote this task as AE for simplicity. 
We believe ABSA is more complete with opinion terms also extracted. Also, the information learned from opinion term extraction could be useful for the other tasks.} 
appearing in a sentence, which is formulated as a sequence tagging problem with the BIO tagging scheme. Specifically, we use five class labels: $Y^{ae} = \{BA, IA, BP, IP, O\}$, indicating the beginning of and inside of an \emph{aspect} term, the beginning of and inside of an \emph{opinion} term, and \emph{other} words, respectively. We also formulate AS as a sequence tagging problem with labels $Y^{as} = \{pos, neg, neu\}$, indicating the token-level \emph{positive}, \emph{negative}, and \emph{neutral} sentiment orientations. Table~\ref{train_example} shows an example of aspect-level training instance with gold AE and AS labels. In aspect-level datasets, only aspect terms get sentiment annotated. Thus, when modeling AS as a sequence tagging problem, we label each token that is part of an aspect term with the sentiment label of the corresponding aspect term. For example, as shown in Table~\ref{train_example}, we label ``\emph{fish}'' as \emph{pos}, and label ``\emph{variety}'', ``\emph{of}'', ``\emph{fish}'' as neg, based on the gold sentiment labels of the two aspect terms ``\emph{fish}'' and ``\emph{varity of fish}'' respectively. Since other tokens do not have AS gold labels, we ignore the predictions on them when computing the training loss for AS.

The AE component $f_{\theta_{ae}}$ is parameterized by $\theta_{ae}$ and outputs $\{\hat{\mathbf{y}}_1^{ae},...,\hat{\mathbf{y}}_n^{ae}\}$. The AS component $f_{\theta_{as}}$ is parameterized by $\theta_{as}$ and outputs $\{\hat{\mathbf{y}}_1^{as},...,\hat{\mathbf{y}}_n^{as}\}$.
The AE and AS encoders consist of $m^{ae}$ and $m^{as}$ layers of CNNs respectively, and they map the shared representations to $\{\mathbf{h}^{ae}_{1}, \mathbf{h}^{ae}_{2}, ..., \mathbf{h}^{ae}_{n}\}$ and $\{\mathbf{h}^{as}_{1}, \mathbf{h}^{as}_{2}, ..., \mathbf{h}^{as}_{n}\}$ respectively. For the AS encoder, we employ an additional self-attention layer on top of the stacked CNNs.
As shown in Figure~\ref{overall_architecture_diagram}, we make $\hat{\mathbf{y}}_i^{ae}$, the outputs from AE available to AS in the self-attention layer, as the sentiment task could benefit from knowing the predictions of opinion terms.  Specifically, the self-attention matrix $\mathbf{A} \in \mathbb{R}^{n \times n}$ is computed as follows:

\begin{small}
\begin{align}
    \text{score}_{ij}^{(i \neq j)} &= (\mathbf{h}_i^{as} \mathbf{W}^{as} (\mathbf{h}_j^{as})^T) \cdot \frac{1}{|i-j|} \cdot P_j^{op}\label{score_fun} \\
    \mathbf{A}_{ij}^{(i \neq j)} &= \frac{\text{exp}(\text{score}_{ij})}{\sum_{k=1}^n \text{exp}(\text{score}_{ik})} 
\end{align}
\end{small}
\noindent where the first term in Eq.(\ref{score_fun}) indicates the semantic relevance between $\mathbf{h}_i^{as}$ and $\mathbf{h}_j^{as}$ with parameter matrix $\mathbf{W}^{as}$, the second term is a distance-relevant factor, which decreases with increasing distance between the $i$th token and the $j$th token, and the third term $P_j^{op}$ denotes the predicted probability that the $j$th token is part of any opinion term. The probability $P_j^{op}$ can be computed by summing the predicted probabilities on opinion-related labels BP and IP in  $\hat{\mathbf{y}}_j^{ae}$. In this way, AS is directly influenced by the predictions of AE. We set the diagonal elements in $\mathbf{A}$ to zeros, as we only consider context words for inferring the sentiment of the target token. The self-attention layer outputs $\mathbf{h}_i^{\prime as} = \sum_{j=1}^n \mathbf{A}_{ij} \mathbf{h}_j^{as}$. In AE, we concatenate the word embedding, the initial shared representation $\mathbf{h}_i^{s(0)}$, and the task-specific representation  $\mathbf{h}_i^{ae}$ as the final representation of the $i$th token. In AS, we concatenate $\mathbf{h}_i^{s(0)}$ and $\mathbf{h}_i^{\prime as}$ as the final representation. For each task, we employ a fully-connected layer with softmax activation as the decoder, which maps the final token representation to probability distribution $\hat{\mathbf{y}}_i^{ae}$ ($\hat{\mathbf{y}}_i^{as}$).

\subsection{Document-Level Tasks}
To address the issue of insufficient aspect-level training data, IMN is able to exploit knowledge from document-level labeled sentiment corpora, which are more readily available. 
We introduce two document-level classification tasks to be jointly trained with AE and AS. One is document-level sentiment classification (DS), which predicts the sentiment towards an input document. The other is document-level domain classification (DD), which predicts the domain label of an input document.

As shown in Figure~\ref{overall_architecture_diagram}, the task-specific operation $f_{\theta_o}$ consists of $m^{o}$ layers of CNNs that map the shared representations $\{\mathbf{h}_1^s, ..., \mathbf{h}_n^s \}$ to $\{\mathbf{h}_1^o, ..., \mathbf{h}_n^o \}$, an attention layer $\text{att}^{o}$, and a decoding layer $\text{dec}^{o}$, where $o \in \{ds, dd\}$ is the task symbol. The attention weight is computed as:
\begin{equation}
    a_i^o = \frac{\text{exp}(\mathbf{h}_i^{o}\mathbf{W}^{o})}{\sum_{k=1}^{n} \text{exp}(\mathbf{h}_k^{o}\mathbf{W}^{o})}
\end{equation}
where $W^{o}$ is a parameter vector. The final document representation is computed as
$\mathbf{h}^{o} = \sum_{i=1}^n a_i^o \mathbf{h}_i^o$. We employ a fully-connected layer with softmax activation as the decoding layer, which maps $\mathbf{h}^o$ to $\hat{\mathbf{y}}^o$.

\subsection{Message Passing Mechanism}
To exploit interactions between different tasks, the message passing mechanism aggregates predictions of different tasks from the previous iteration, and uses this knowledge to update the shared latent vectors $\{\mathbf{h}^s_{1}, ..., \mathbf{h}^s_{n}\}$ at the current iteration. 
Specifically, the message passing mechanism integrates knowledge from $\hat{\mathbf{y}}^{ae}_i$, $\hat{\mathbf{y}}^{as}_i$, $\hat{\mathbf{y}}^{ds}$, $a_i^{ds}$, and $a_i^{dd}$ computed on an input \{$x_1, ..., x_n$\}, and the shared hidden vector $\mathbf{h}^s_{i}$ is updated as follows:
\begin{equation}
\resizebox{.87\hsize}{!}{$
\begin{split}
    \mathbf{h}^{s(t)}_i= 
    &f_{\theta_{re}}(\mathbf{h}_i^{s(t-1)}: \hat{\mathbf{y}}_i^{ae(t-1)}: \hat{\mathbf{y}}_i^{as(t-1)}:\\ &\hat{\mathbf{y}}^{ds(t-1)}: a_i^{ds(t-1)}: a_i^{dd(t-1)})
\end{split}
$}\label{update_hidden}
\end{equation}
where $t>0$ and $[:]$ denotes the concatenation operation.
We employ a fully-connected layer with ReLu activation as the re-encoding function $f_{\theta_{re}}$. 
To update the shared representations, we incorporate $\hat{\mathbf{y}}_i^{ae(t-1)}$ and $\hat{\mathbf{y}}_i^{as(t-1)}$, the outputs of AE and AS from the previous iteration, such that these information are available for both tasks in current round of computation.
We also incorporate information from DS and DD. $\hat{\mathbf{y}}^{ds}$ indicates the overall sentiment of the input sequence, which could be helpful for AS. The attention weights $a_i^{ds}$ and $a_i^{dd}$ generated by DS and DD respectively reflect how sentiment-relevant and domain-relevant the $i$th token is. A token that is more sentiment-relevant or domain-relevant is more likely to be an opinion word or aspect word. This information is useful for the aspect-level tasks.

\subsection{Learning}
Instances for aspect-level problems only have aspect-level labels while instances for document-level problems only have document labels.  IMN is trained on aspect-level and document-level instances alternately. 

When trained on aspect-level instances, the loss function is as follows:
\begin{equation}
\resizebox{.87\hsize}{!}{$
    \begin{split}
     L_{a}(\theta_{s}, &\theta_{ae}, \theta_{as}, \theta_{ds}, \theta_{dd}, \theta_{re}) = \frac{1}{N_a}\sum\limits_{i=1}^{N_a} \frac{1}{n_i} \sum\limits_{j=1}^{n_i} (\\ &l(\mathbf{y}_{i,j}^{ae}, \hat{\mathbf{y}}_{i,j}^{ae(T)})+l(\mathbf{y}_{i,j}^{as}, \hat{\mathbf{y}}_{i,j}^{as(T)}))
    \end{split}
$}\label{aspect_loss}
\end{equation}
where $T$ denotes the maximum number of iterations in the message passing mechanism, $N_a$ denotes the total number of aspect-level training instances, $n_i$ denotes the number of tokens contained in the $i$th training instance, and $\mathbf{y}_{i,j}^{ae}$ ($\mathbf{y}_{i,j}^{as}$) denotes the one-hot encoding of the gold label for AE (AS). $l$ is the cross-entropy loss applied to each token. In aspect-level datasets, only aspect terms have sentiment annotations. We label each token that is part of any aspect term with the sentiment of the corresponding aspect term. During model training, we only consider AS predictions on these aspect term-related tokens for computing the AS loss and ignore the sentiments predicted on other tokens\footnote{Let $l(\mathbf{y}_{i,j}^{as}, \hat{\mathbf{y}}_{i,j}^{as(T)}) = 0$ in Eq.(\ref{aspect_loss}) if $\mathbf{y}_{i,j}^{ae}$ is not BA or IA}.

When trained on document-level instances, we minimize the following loss:
\begin{equation}
\resizebox{.7\hsize}{!}{$
\begin{split}
   L_d(\theta_s, \theta_{ds}, \theta_{dd}) &= \frac{1}{N_{ds}} \sum_{i=1}^{N_{ds}} l( \mathbf{y}_i^{ds}, \hat{\mathbf{y}}_i^{ds})\\
    &+ \frac{1}{N_{dd}} \sum_{i=1}^{N_{dd}} l( \mathbf{y}_i^{dd}, \hat{\mathbf{y}}_i^{dd})
\end{split}
$}
\end{equation}
\noindent where $N_{ds}$ and $N_{dd}$ denote the number of training instances for DS and DD respectively, and $\mathbf{y}_i^{ds}$ and $\mathbf{y}_i^{dd}$ denote the one-hot encoding of the gold label. Message passing iterations are not used when training document-level instances.

\begin{algorithm}[t]
\caption{Pseudocode for training IMN }
\begin{algorithmic} 
\REQUIRE $D^{a} = \{(x^a_i, y^{ae}_i, y^{as}_i)_{i=1}^{N_a}\}$, $D^{ds}=\{(x_i^{ds}, y_i^{ds})_{i=1}^{N_{ds}}\}$ and $D^{dd}=\{(x_i^{dd}, y_i^{dd})_{i=1}^{N_{dd}}\}$
\REQUIRE $\text{Integer} \quad r>0$
\medskip

\FOR{$e \in [1, \text{\emph{max-pretrain-epochs}}]$}
    \FOR{minibatch $B^{ds}$, $B^{dd}$ in 
     $D^{ds}$, $D^{dd}$}
     \STATE compute $L_d$ based on $B^{ds}$ and $B^{dd}$
     \STATE update $\theta_{s}$, $\theta_{ds}$, $\theta_{dd}$
     \ENDFOR
\ENDFOR
\medskip

\FOR{$e \in [1, \text{\emph{max-epochs}}]$}
    \FOR{$b \in [1, \text{\emph{batches-per-epoch}}]$}
    \STATE sample $B^{a}$ from $D^{a}$
    \STATE compute $L_a$ based on $B^{a}$
    \STATE update $\theta_{s}$, $\theta_{ae}$, $\theta_{as}$, $\theta_{re}$
    \IF{$b$ is divisible by $r$}
    \STATE sample $B^{ds}$, $B^{dd}$ from $D^{ds}$, $D^{dd}$
     \STATE compute $L_d$ based on $B^{ds}$ and $B^{dd}$
     \STATE update $\theta_{s}$, $\theta_{ds}$, $\theta_{dd}$
     \ENDIF
    \ENDFOR
\ENDFOR
\end{algorithmic}\label{algo}
\end{algorithm}

For learning, we first pretrain the network on the document-level instances (minimize $L_d$) for a few epochs, such that DS and DD can make reasonable predictions. Then the network is trained on aspect-level instances and document-level instances alternately with ratio $r$, to minimize $L_a$ and $L_d$. The overall training process is given in Algorithm~\ref{algo}. $D^{a}$, $D^{ds}$, and $D^{dd}$ denote the aspect-level training set and the training sets for DS, DD respectively. $D^{ds}$ and $D^{a}$ are from similar domains. $D^{dd}$ contains review documents from at least two domains with $y_i^{ds}$ denoting the domain label, where one of the domains is similar to the domains of $D^{a}$ and $D^{ds}$. In this way, linguistic knowledge can be transferred from DS and DD to AE and AS, as they are semantically relevant. We fix $\theta_{ds}$ and $\theta_{dd}$ when updating parameters for $L_a$, since we do not want them to be affected by the small number of aspect-level training instances.

\renewcommand{\arraystretch}{1.1}
\begin{table}[t]
\centering
\small
\scalebox{0.9}{
\begin{tabular}{llcccc}
\toprule 
&\multirow{ 2}{*}{Datasets} & \multicolumn{2}{c}{Train} &  \multicolumn{2}{c}{Test}\\\cline{3-6}
&& aspect & opinion & aspect &opinion\\\hline
D1&Restaurant14 &3699 &3484 &1134 &1008\\
D2&Laptop14 &2373 &2504 &654 &674 \\
D3&Restaurant15 &1199 &1210 &542 &510\\
\bottomrule
\end{tabular}}
\caption{Dataset statistics with numbers of aspect terms and opinion terms}\label{data}
\end{table}

\section{Experiments}
\subsection{Experimental Settings}
\textbf{Datasets.} Table~\ref{data} shows the statistics of the aspect-level datasets. We run experiments on three benchmark datasets, taken from SemEval2014~\citep{Pontiki:14} and SemEval 2015~\citep{Pontiki:15}. The opinion terms are annotated by~\citet{Wang:16b}. We use two document-level datasets from~\citep{He:18}. One is from the Yelp restaurant domain, and the other is from the Amazon electronics domain. Each contains 30k instances with exactly balanced class labels of $pos$, $neg$, and $neu$. We use the concatenation of the two datasets with domain labels as $D^{dd}$. We use the Yelp dataset as $D^{ds}$ when $D^{a}$ is either D1 or D3, and use the electronics dataset as $D^{ds}$ when $D^{a}$ is D2.
\medskip

\noindent\textbf{Network details.} We adopt the multi-layer-CNN structure from~\citep{hu:18} as the CNN-based encoders in our proposed network. See Appendix~\ref{sec:enc} for implementation details.  
For word embedding initialization, we concatenate a general-purpose embedding matrix and a domain-specific embedding matrix\footnote{For DD, we only look at the general-purpose embeddings by masking out the domain-specific embeddings.} following ~\citep{hu:18}. We adopt their released domain-specific embeddings for restaurant and laptop domains with 100 dimensions, which are trained on a large domain-specific corpus using fastText. The general-purpose embeddings are pre-trained Glove vectors~\citep{Pennington:14} with 300 dimensions. 

One set of important hyper-parameters are the number of CNN layers in the shared encoder and the task-specific encoders. To decide the values of $m^{s}$, $m^{ae}$, $m^{as}$, $m^{ds}$, $m^{dd}$, we first investigate how many layers of CNNs would work well for each of the task when training it alone. We denote $c^{o}$ as the optimal number of CNN layers in this case, where $o \in \{ae, as, ds, dd\}$ is the task indicator. We perform AE, AS separately on the training set of D1, and perform DS, DD separately on the document-level restaurant corpus. Cross-validation is used for selecting $c^{o}$, which yields 4, 2, 2, 2 for $c^{ae}$, $c^{as}$, $c^{ds}$, $c^{dd}$. Based on this observation, we made $m^{s}$, $m^{ae}$, $m^{as}$, $m^{ds}$, $m^{dd}$ equals to 2, 2, 0, 0, 0 respectively, such that $m^{s}+m^{o}=c^{o}$. Note that there are other configurations satisfying the requirement, for example, $m^{s}$, $m^{ae}$, $m^{as}$, $m^{ds}$, $m^{dd}$ equals to 1, 3, 1, 1, 1. we select our setting as it involves the smallest set of parameters. 

We tune the maximum number of iterations $T$ in the message passing mechanism by training IMN$^{-d}$ via cross validation on D1. It is set to 2. With $T$ fixed as 2, we then tune $r$ by training IMN via cross validation on D1 and the relevant document-level datasets. It is set to 2 as well. 

We use Adam optimizer with learning rate set to $10^{-4}$, and we set batch size to 32. Learning rate and batch size are set to conventional values without specific tuning for our task. 

At training phase, we randomly sample 20\% of the training data from the aspect-level dataset as the development set and only use the remaining 80\% for training. We train the model for a fix number of epoches, and save the model at the epoch with the best F1-I score on the development set for evaluation. 
\medskip

\noindent\textbf{Evaluation metrics.} 
During testing, we extract aspect (opinion) terms, and predict the sentiment for each extracted aspect term based on $\hat{\mathbf{y}}^{ae(T)}$ and $\hat{\mathbf{y}}^{as(T)}$. Since the extracted aspect term may consist of multiple tokens and the sentiment predictions on them could be inconsistent in AS, we only output the sentiment label of the first token as the predicted sentiment for any extracted aspect term. 

We employ five metrics for evaluation, where two measure the AE performance, two measure the AS performance, and one measures the overall performance. 
Following existing works for AE~\citep{Wang:17, hu:18}, we use F1 to measure the performance of aspect term extraction and opinion term extraction, which are denoted as \textbf{F1-a} and \textbf{F1-o} respectively. 
Following existing works for AS~\citep{Chen:17, He:18}, we adopt accuracy and macro-F1 to measure the performance of AS. We denote them as \textbf{acc-s} and \textbf{F1-s}. Since we are solving the integrated task without assuming that gold aspect terms are given, the two metrics are computed based on the correctly extracted aspect terms from AE.  
We compute the F1 score of the integrated task denoted as \textbf{F1-I} for measuring the overall performance. To compute F1-I, an extracted aspect term is taken as correct only when both the span and the sentiment are correctly identified. When computing F1-a, we consider all aspect terms, while when computing acc-s, F1-s, and F1-I, we ignore aspect terms with \emph{conflict} sentiment labels.

\renewcommand{\arraystretch}{1.0}
\begin{table*}[t]
\centering
\small
\scalebox{0.93}{
\begin{tabular}{l|c|ccccc|cc|cccc}
&\rotatebox{90}{Methods}&\rotatebox{90}{CMLA-ALSTM}&\rotatebox{90}{CMLA-dTrans}&\rotatebox{90}{DECNN-ALSTM}&\rotatebox{90}{DECNN-dTrans}&\rotatebox{90}{PIPELINE}&\rotatebox{90}{MNN}&\rotatebox{90}{INABSA}&\rotatebox{90}{IMN$^{-d}$ wo DE}&\rotatebox{90}{IMN$^{-d}$}&\rotatebox{90}{IMN wo DE}&\rotatebox{90}{IMN}\\
\hline

\multirow{5}{*}{\rotatebox{90}{D1}}&F1-a &82.45 &82.45 &83.94 &83.94 &83.94 &83.05 &83.92 &83.95 &\bf{84.01} &83.50 &83.33\\
&F1-o &82.67 &82.67 &85.60 &85.60 &85.60 &84.55 &84.97 &85.21 &\bf{85.64} &84.62 &85.61\\
&acc-s &77.46 &79.58 &77.79 &80.04 &79.56 &77.17 &79.68 &79.65 &81.56$^*$ &83.17$^*$ &\bf{83.89}$^*$\\
&F1-s &68.70 &72.23 &68.50 &73.31 &69.59 &68.45 &68.38 &69.32 &71.90 &73.44 &\bf{75.66}\\
&F1-I &63.87 &65.34 &65.26 &67.25 &66.53 &63.87 &66.60 &66.96 &68.32$^*$ &69.11$^*$ &\bf{69.54}$^*$\\\hline 

\multirow{5}{*}{\rotatebox{90}{D2}}&F1-a &76.80 &76.80 &78.38 &78.38 &78.38 &76.94 &77.34 &76.96 &\bf{78.46} &76.87 &77.96\\
&F1-o &77.33 &77.33 &\bf{78.81} &\bf{78.81} &\bf{78.81} &77.77 &76.62 &76.85 &78.14 &77.04 &77.51\\
&acc-s &70.25 &72.38 &70.46 &73.10 &72.29 &70.40 &72.30 &72.89 &73.21 &74.31$^*$ &\bf{75.36}$^*$\\
&F1-s &66.67 &69.52 &66.78 &70.63 &68.12 &65.98 &68.24 &67.26 &69.92 &70.76 &\bf{72.02}$^*$\\
&F1-I &53.68 &55.56 &55.05 &56.60 &56.02 &53.80 &55.88 &56.25 &57.66$^*$ &57.04$^*$ &\bf{58.37}$^*$\\\hline

\multirow{5}{*}{\rotatebox{90}{D3}}&F1-a &68.55 &68.55 &68.32 &68.32 &68.32 &\bf{70.24} &69.40 &69.23 &69.80 &68.23 &70.04\\
&F1-o &71.07 &71.07 &71.22 &71.22 &71.22 &69.38 &71.43 &68.39 &\bf{72.11}$^*$ &70.09 &71.94\\
&acc-s &81.03 &82.27 &80.32 &82.65 &82.27 &80.79 &82.56 &81.64 &83.38 &\bf{85.90}$^*$ &85.64$^*$\\
&F1-s &58.91 &66.45 &57.25 &69.58 &59.53 &57.90 &58.81 &57.51 &60.65 &71.67$^*$ &\bf{71.76}$^*$\\
&F1-I &54.79 &56.09 &55.10 &56.28 &55.96 &56.57 &57.38 &56.80 &57.91$^*$ &58.82$^*$ &\bf{59.18}$^*$\\\hline
\end{tabular}}
\caption{Model comparison. Average results over 5 runs with random initialization are reported. $^*$ indicates the proposed method is significantly better than the other baselines ($p<0.05$) based on one-tailed unpaired t-test. }\label{main results}
\end{table*}

\subsection{Models under Comparison}
\textbf{Pipeline approach. } 
We select two top-performing models from prior works for each of AE and AS, to construct $2 \times 2$ pipeline baselines.
For AE, we use CMLA~\citep{Wang:17} and DECNN~\citep{hu:18}. CMLA was proposed to perform co-extraction of aspect and opinion terms by modeling their inter-dependencies. DECNN is the state-of-the-art model for AE. It utilizes a multi-layer CNN structure with both general-purpose and domain-specific embeddings. We use the same structure as encoders in IMN.
For AS, we use ATAE-LSTM (denoted as ALSTM for short)~\citep{Wang:16} and the model from~\citep{He:18} which we denote as dTrans. ALSTM is a representative work with an attention-based LSTM structure. We compare with dTrans as it also utilizes knowledge from document corpora for improving AS performance, which achieves state-of-the-art results.  Thus, we compare with the following pipeline methods: \textbf{CMLA-ALSTM}, \textbf{CMLA-dTrans}, \textbf{DECNN-ALSTM}, and \textbf{DECNN-dTrans}. We also compare with the pipeline setting of IMN, which trains AE and AS independently (i.e., without parameter sharing, information passing, and document-level corpora). We denote it as \textbf{PIPELINE}. The network structure for AE in PIPELINE is the same as DECNN. During testing of all methods, we perform AE in the first step, and then generate AS predictions on the correctly extracted aspect terms. 
\smallskip

\noindent\textbf{Integrated Approach. } We compare with two recently proposed methods that have achieved state-of-the-art results among integrated approaches: \textbf{MNN}~\citep{Wang:18b} and the model from~\citep{Li:19} which we denote as \textbf{INABSA} (integrated network for ABSA). Both methods model the overall task as a sequence tagging problem with a unified tagging scheme. Since during testing, IMN only outputs the sentiment on the first token of an extracted aspect term to avoid sentiment inconsistency, to enable fair comparison, we also perform this operation on MNN and INABSA. We also show results for a version of IMN that does not use document-level corpora, denoted as IMN$^{-d}$. The structure of IMN$^{-d}$ is shown as the solid lines in Figure~\ref{overall_architecture_diagram}. It omits the information $\hat{\mathbf{y}}^{ds}$, $a_i^{ds}$, and $a_i^{dd}$ propagated from the document-level tasks in Eq.(\ref{update_hidden}).
\medskip

\subsection{Results and Analysis}
\textbf{Main results.} 
Table~\ref{main results} shows the comparison results.
Note that IMN performs co-extraction of aspect and opinion terms in AE, which utilizes additional opinion term labels during training, while the baseline methods except CMLA do not consider this information in their original models. To enable fair comparison, we slightly modify those baselines to perform co-extraction as well, with opinion term labels provided. Further details on model comparison are provided in Appendix~\ref{model_comparison}.

 From Table~\ref{main results}, we observe that IMN$^{-d}$ is able to significantly outperform other baselines on F1-I. IMN further boosts the performance and outperforms the best F1-I results from the baselines by 2.29\%, 1.77\%, and 2.61\%  on D1, D2, and D3.
 Specifically, for AE (F1-a and F1-o), IMN$^{-d}$ performs the best in most cases.
 For AS (acc-s and F1-s), IMN outperforms other methods by large margins. PIPELINE, IMN$^{-d}$, and the pipeline methods with dTrans also perform reasonably well on this task, outperforming other baselines by moderate margins. All these models utilize knowledge from larger corpora by either joint training of document-level tasks or using domain-specific embeddings. This suggests that domain-specific knowledge is very helpful, and both joint training and domain-specific embeddings are effective ways to transfer such knowledge.

\renewcommand{\arraystretch}{1.2}
\begin{table}[t]
\centering
\small
\scalebox{0.9}{
\begin{tabular}{lccc}
\toprule 
Model variants&D1&D2&D3\\\hline
Vanilla model &66.66 &55.63 &56.24\\
+Opinion transmission  &66.98 &56.03 &56.65\\
+Message passing-a (IMN$^{-d}$) &68.32 &57.66 &57.91\\
+DS &68.48 &57.86 &58.03\\
+DD &68.65 &57.50 &58.26\\
+Message passing-d  (IMN) &69.54 &58.37 &59.18\\
\bottomrule
\end{tabular}}
\caption{F1-I scores of different model variants. Average results over 5 runs are reported.}\label{ablation test}
\end{table}

\renewcommand{\arraystretch}{1.2}
\begin{table*}[t]
\centering
\small
\scalebox{0.8}{
\begin{tabular}{l|l|l|l|l|l|l}
\toprule 
\multirow{ 2}{*}{Examples} & \multicolumn{2}{c}{PIPELINE} & \multicolumn{2}{c}{INABSA} & \multicolumn{2}{c}{IMN}\\\cline{2-7}
&Opinion &Aspect &Opinion &Aspect &Opinion &Aspect \\\hline
1. \makecell[l]{\textcolor{blue}{\emph{Strong}} [\textcolor{red}{build}]$_{\text{pos}}$ though which really \\ adds to its [\textcolor{red}{durability}]$_{\text{pos}}$.} &Strong &[durability]$_{\text{pos}}$ &Strong &[durability]$_{\text{pos}}$ &Strong &\makecell[l]{[build]$_{\text{pos}}$,  [durability]$_{\text{pos}}$}\\\hline
2. \makecell[l]{Curioni's Pizza has been around since \\ the 1920's} & None &[Pizza]$_{\text{neu}}$ &None &[Pizza]$_{\text{pos}}$ &None &None\\\hline
3. The [\textcolor{red}{battery}]$_{\text{pos}}$ is \textcolor{blue}{\emph{longer}}&longer &[battery]$_{\text{neg}}$ &longer &[battery]$_{\text{neg}}$ &longer &[battery]$_{\text{pos}}$\\\hline
4. The [\textcolor{red}{potato balls}]$_{\text{pos}}$ were not \textcolor{blue}{\emph{dry}} at all &dry &[potato balls]$_{\text{neg}}$ &dry &[potato balls]$_{\text{neg}}$ &dry &[potato balls]$_{\text{pos}}$\\\hline
5. \makecell[l]{That's a \textcolor{blue}{\emph{good}} thing, but it's made \\ from [\textcolor{red}{aluminum}]$_{\text{neg}}$ that \textcolor{blue}{\emph{scratches easily}}}.&\makecell[l]{good, \\ easily} &[aluminum]$_{\text{pos}}$ &\makecell[l]{good, \\ easily} &[aluminum]$_{\text{pos}}$ &\makecell[l]{good, \\ scratches easily} &[aluminum]$_{\text{neg}}$\\\hline
\bottomrule
\end{tabular}}
\caption{Case analysis. The ``Examples'' column contains instances with gold labels. 'The ``opinion'' and ``aspect'' columns present the opinion terms and aspect terms with sentiments, generated by the corresponding model.}\label{case_study}
\end{table*}

\renewcommand{\arraystretch}{1.2}
\begin{table}[t]
\centering
\small
\scalebox{0.9}{
\begin{tabular}{lcccccc}
\toprule 
$T$&0&1&2&3&4&5\\\hline
D1 &66.98 &67.97 &\bf{68.32} &68.03 &68.11 &68.26\\
D2 &56.03 &57.14 &57.66 &\bf{57.82} &57.78 &57.33\\
D3 &56.65 &57.60 &\bf{57.91} &57.66 &57.41 &57.48\\

\bottomrule
\end{tabular}}
\caption{F1 scores with different $T$ values using IMN$^{-d}$. Average results over 5 runs are reported.}\label{vary T}
\end{table}

 We also show the results of IMN$^{-d}$ and IMN when only the general-purpose embeddings (without domain-specific embeddings) are used for initialization. They are denoted as IMN$^{-d}$/IMN wo DE. IMN wo DE performs only marginally below IMN. This indicates that the knowledge captured by domain-specific embeddings could be similar to that captured by joint training of the document-level tasks. IMN$^{-d}$ is more affected without domain-specific embeddings, while it still outperforms all other baselines except DECNN-dTrans. DECNN-dTrans is a very strong baseline as it exploits additional knowledge from larger corpora for both tasks. IMN$^{-d}$ wo DE is competitive with DECNN-dTrans even without utilizing additional knowledge, which suggests the effectiveness of the proposed network structure. 
\medskip

\noindent\textbf{Ablation study.} To investigate the impact of different components, we start with a vanilla model which consists of $f_{\theta_s}$, $f_{\theta_{ae}}$, and $f_{\theta_{as}}$ only without any informative message passing, and add other components one at a time. Table~\ref{ablation test} shows the results of different model variants. +Opinion transmission denotes the operation of providing additional information $P_j^{op}$ to the self-attention layer as shown in Eq.(\ref{score_fun}). +Message passing-a denotes propagating the outputs from aspect-level tasks only at each message passing iteration. +DS and +DD denote adding DS and DD with parameter sharing only.  +Message passing-d denotes involving the document-level information for message passing.  We observe that +Message passing-a and +Message passing-d contribute to the performance gains the most, which demonstrates the effectiveness of the proposed message passing mechanism. We also observe that simply adding document-level tasks (+DS/DD) with parameter sharing only marginally improves the performance of IMN$^{-d}$. This again indicates that domain-specific knowledge has already been captured by domain embeddings, while knowledge obtained from DD and DS via parameter sharing could be redundant in this case. However, +Message passing-d is still helpful with considerable performance gains, showing that aspect-level tasks can benefit from knowing predictions of the relevant document-level tasks. 
\medskip

\noindent\textbf{Impact of $T$.}
We have demonstrated the effectiveness of the message passing mechanism. Here, we investigate the impact of the maximum number of iterations $T$. Table~\ref{vary T} shows the change of F1-I on the test sets as $T$ increases. We find that convergence is quickly achieved within two or three iterations, and further iterations do not provide considerable performance improvement. 
\medskip

\noindent\textbf{Case study.}
To better understand in which conditions the proposed method helps, we examine the instances that are misclassified by PIPELINE and INABSA, but correctly classified by IMN. 

For aspect extraction, we find the message passing mechanism is particularly helpful in two scenarios. First, it helps to better recognize uncommon aspect terms by utilizing information from the opinion contexts. As shown in example 1 in Table~\ref{case_study}, PIPELINE and INABSA fail to recognize ``build'' as it is an uncommon aspect term in the training set while IMN is able to correctly recognize it. We find that when no message passing iteration is performed, IMN also fails to recognize ``build''. However, when we analyze the predicted sentiment distribution on each token in the sentence, we find that except ``durability'', only ``build'' has a strong positive sentiment, while the sentiment distributions on the other tokens are more uniform. This is an indicator that ``build'' is also an aspect term. IMN is able to aggregate such knowledge with the message passing mechanism, such that it is able to correctly recognize ``build'' in later iterations. 
Due to the same reason, the message passing mechanism also helps to avoid extracting terms on which no opinion is expressed. As observed in example 2, both PIPELINE and INABSA extract ``Pizza''. However, since no opinion is expressed in the given sentence, ``Pizza'' should not be considered as an aspect term. IMN avoids extracting this kind of terms by aggregating knowledge from opinion prediction and sentiment prediction. 

For aspect-level sentiment, since IMN is trained on larger document-level labeled corpora with balanced sentiment classes, in general it better captures the meaning of domain-specific opinion words (example 3), better captures sentiments of complex expressions such as negation (example 4), and better recognizes minor sentiment classes in the aspect-level datasets (negative and neutral in our cases). In addition, we find that knowledge propagated by the document-level tasks through message passing is helpful.  
For example, the sentiment-relevant attention weights are helpful for recognizing uncommon opinion words, and which further help on correctly predicting the sentiments of the aspect terms. 
As observed in example 5, PIPELINE and INABSA are unable to recognize ``scratches easily'' as the opinion term, and they also make wrong sentiment prediction on the aspect term ``aluminum''. IMN learns that ``scratches'' is sentiment-relevant through knowledge from the sentiment-relevant attention weights aggregated via previous iterations of message passing, and is thus able to extract ``scratches easily''. Since the opinion predictions from AE are sent to the self-attention layer in the AS component, correct opinion predictions further help to infer the correct sentiment towards ``aluminum''.

\section{Conclusion}

We propose an interactive multi-task learning network IMN for jointly learning aspect and opinion term co-extraction, and aspect-level sentiment classification. The proposed IMN introduces a novel message passing mechanism that allows informative interactions between tasks, enabling the correlation to be better exploited. In addition, IMN is able to learn from multiple training data sources, allowing fine-grained token-level tasks to benefit from document-level labeled corpora. The proposed architecture can potentially be applied to similar tasks such as relation extraction, semantic role labeling, etc. 

\section*{Acknowledgments}
This research is supported by the National Research Foundation Singapore under its AI Singapore Programme grant AISG-RP-2018-006.

\bibliography{acl2019}
\bibliographystyle{acl_natbib}

\appendix
\section{Implementation Details}
\label{sec:enc}

\textbf{CNN-based Encoder}

\noindent We adopt the multi-layer-CNN structure from~\citep{hu:18} as the CNN-based encoders for both the shared CNNs and the task-specific ones in the proposed network. 
Each CNN layer has many 1D-convolution filters, and each filter has a fixed kernel size $k=2c+1$, such that each filter performs convolution operation on a window of $k$ word representations, and compute the representation for the $i$th word along with 2$c$ nearby words in its context.

Following the settings in the original paper, the first CNN layer in the shared encoder has 128 filters with kernel sizes $k=3$ and 128 filters with kernel sizes $k=5$. The other CNN layers in the shared encoder and the CNN layers in each task-specific encoder have 256 filters with kernel sizes $k=5$ per layer. ReLu is used as the activation function for each CNN layer. Dropout with $p=0.5$ is employed after the embedding layer and each CNN layer. 

\medskip

\noindent\textbf{Opinion Transmission}

\noindent To alleviate the problem of unreliable predictions of opinion labels in the early stage of training, we adopt scheduled sampling for opinion transmission at training phase. We send gold opinion labels rather than the predicted ones generated by AE to AS in the probability of $\epsilon_i$. The probability $\epsilon_i$ depends on the number of epochs $i$ during training, for which we employ an inverse sigmoid decay $\epsilon_i = 5/(5+\exp(i/5))$.

\renewcommand{\arraystretch}{1.1}
\begin{table*}[t]
\centering
\small
\scalebox{0.9}{
\begin{tabular}{l|cccc|cccc|cccc}
\toprule
\multirow{ 2}{*}{Methods} & \multicolumn{4}{c}{D1} &  \multicolumn{4}{c}{D2} & \multicolumn{4}{c}{D3}\\\cline{2-13}
& F1-a  & acc-s & F1-s & F1-I & F1-a  & acc-s & F1-s & F1-I & F1-a  & acc-s & F1-s & F1-I\\\hline
DECNN-ALSTM &83.33 &77.63 &70.09 &64.32 &\bf{80.28} &69.98 &66.20 &55.92 &68.72 &79.22 &54.40 &54.22\\
DECNN-dTrans &83.33 &79.45 &73.08 &66.15 &\bf{80.28} &71.51 &68.03 &57.28 &68.72 &82.09 &68.35 &56.08\\
PIPELINE &83.33 &79.39 &69.45 &65.96 &\bf{80.28} &72.12 &68.56 &57.29 &68.72 &81.85 &58.74 &56.04\\\hline
MNN  &83.20 &77.57 &68.19 &64.26 &76.33 &70.62 &65.44 &53.77 &69.29 &80.86 &55.45 &55.93\\
INABSA &83.12 &79.06 &68.77 &65.94 &77.67 &71.72 &68.36 &55.95 &68.79 &80.96 &57.10 &55.45\\\hline
IMN$^{-d}$&\bf{83.89} &80.69 &72.09 &67.27$^*$ &78.43 &72.49 &69.71 &57.13 &\bf{70.35}$^*$ &81.86 &56.88 &57.86$^*$ \\
IMN &83.04 &\bf{83.05}$^*$ &\bf{73.30} &\bf{68.71}$^*$ &77.69 &\bf{75.12}$^*$ &\bf{71.35}$^*$ &\bf{58.04}$^*$ &69.25 &\bf{84.53}$^*$ &\bf{70.85}$^*$ &\bf{58.18}$^*$\\

\bottomrule
\end{tabular}}
\caption{Model comparison in a setting without opinion term labels. Average results over 5 runs with random initialization are reported. $^*$ indicates the proposed method is significantly better than the other baselines ($p<0.05$) based on one-tailed unpaired t-test. }\label{main results_wo_op}
\end{table*}

\section{Model Comparison Details}
\label{model_comparison}
For CMLA\footnote{\url{https://github.com/happywwy/Coupled-Multi-layer-Attentions}},  ALSTM\footnote{\url{https://www.wangyequan.com/publications/}}, dTrans\footnote{\url{https://github.com/ruidan/Aspect-level-sentiment}}, and INABSA\footnote{\url{https://github.com/lixin4ever/E2E-TBSA}}, we use the officially released source codes for experiments. For MNN, we re-implement the model following the descriptions in the paper as the source code is not available. We run each baseline multiple times with random initializations and save their predicted results. We use an unified evaluation script for measuring the outputs from different baselines as well as the proposed method.

The proposed IMN performs co-extraction of aspect terms and opinion terms in AE, which utilizes additional opinion term labels during model training. In the baselines, the two integrated methods MNN and INABSA, and the pipeline methods with DECNN as the AE component do not take take opinion information during training. To make fair comparison, we add labels $\{$BP, IP$\}$ to the original label sets of MNN, INABSA, and DECNN, indicating the beginning of and inside of an opinion term. We train those models on training sets with both aspect and opinion term labels to perform co-extraction as well. In addition, for pipeline methods, we also make the gold opinion terms available to the AS models (ALSTM and dTrans) during training. To make ALSTM and dTrans utilize the opinion label information, we modify their attention layer to assign higher weights to tokens that are more likely to be part of an opinion term. This is reasonable since the objective of the attention mechanism in an AS model is to find the relevant opinion context.  The attention weight of the $i$th token before applying softmax normalization in an input sentence is modified as:
\begin{equation}
    a_i^{\prime} = a_i * P_i^{op} 
\end{equation}
where $a_i$ denotes the attention weight computed by the original attention layer, $p_i^{op}$ denotes the probability that the $i$th token belongs to any opinion term. $a_i^{\prime}$ denotes the modified attention weights. At the training phase, since the gold opinion terms are provided, $p_i^{op}=1$ for the tokens that are part of the gold opinion terms, while $p_i^{op}=0$ for the other tokens. At the testing phase, $p_i^{op}$ is computed based on the predictions from the AE model in the pipeline method. It is computed by summing up the predicted probabilities on the opinion-related labels BP and IP for the $i$th token.

We also present the comparison results in a setting without using opinion term labels in Table~\ref{main results_wo_op}\footnote{We exclude the results of the pipeline methods with CMLA, as CMLA relies on opinion term labels during training. It is difficult to modify it.}. In this setting, we modify the proposed IMN and IMN$^{-d}$ to recognize aspect terms only in AE. The opinion transmission operation, which sends the opinion term predictions from AE to AS, is omitted as well. 

Both IMN$^{-d}$ and IMN still significantly outperform other baselines in most cases under this setting. In addition, when compare the results in Table~\ref{main results_wo_op} and Table~\ref{main results}, we observe that IMN$^{-d}$ and IMN consistently yield better F1-I scores on all datasets in Table~\ref{main results}, when opinion term extraction is also considered. Consistent improvements are not observed in other baseline methods when trained with opinion term labels. These findings suggest that knowledge obtained from learning opinion term extraction is indeed beneficial, however, a carefully-designed network structure is needed to utilize such information. IMN is designed to exploit task correlations by explicitly modeling interactions between tasks, and thus it better integrates knowledge obtained from training different tasks.

\end{document}